\documentclass[11pt]{article}

\usepackage[utf8]{inputenc}
\usepackage[T1]{fontenc}
\usepackage{mathptmx}
\usepackage[scaled=0.9]{helvet}
\usepackage[letterpaper,margin=1in]{geometry}
\usepackage{graphicx}
\graphicspath{{./figures/}{../assets/figures/}{./}}
\usepackage{booktabs}
\usepackage{amsmath}
\usepackage{xcolor}
\usepackage{microtype}
\usepackage{caption}
\usepackage[hidelinks]{hyperref}
\usepackage{url}

\usepackage{authblk}
\usepackage{longtable}
\usepackage{array}
\usepackage{enumitem}
\usepackage{tikz}
\usetikzlibrary{arrows.meta,positioning,fit,backgrounds}

\title{\bfseries Manac\'a-1B-Instruct-2: Reproducible Safety Alignment for a\\
Native Brazilian-Portuguese LLM\\[0.3em]
\large Three Negative DPO Attempts, a Three-Iteration Safety-SFT, a Controlled\\
Composition Scan, and a Silent Failure Mode of the LLM Judge in Safety Evaluation}

\author[1]{Bruno Leonardo Santos Menezes}
\author[1]{Carlos Leonardo Souza Cardoso}
\author[1]{Fabio Andr\'e Machado Porto}
\affil[1]{National Laboratory for Scientific Computing (LNCC), Institute of Artificial Intelligence\\
Petr\'opolis, Brazil, \texttt{\{brunolsm, cardoso, fporto\}@lncc.br}}
\date{LNCC $\times$ NII/LLM-jp Scientific Cooperation \\ \today}

\begin{document}
\maketitle

\begin{center}
\small License: code under CC BY 4.0; weights under CC BY-NC 4.0 (see \S\ref{sec:license}) \\
Code: \url{https://github.com/Instituto-IA-LNCC/manaca-1b-instruct-2} \\
Weights: \url{https://huggingface.co/AIInstituteLNCC/manaca-1b-instruct-2}
\end{center}
\vspace{0.5em}

\begin{abstract}
\noindent
Brazilian Portuguese remains underserved by open-source language models, and the
few that are safety-aligned rarely document what did \emph{not} work or validate
the comparison with statistical significance. We present Manac\'a-1B-Instruct-2,
the second generation of the post-training alignment of Manac\'a-1B (1.72 billion
parameters, trained from scratch for PT-BR), with an SFT corpus of 291,223
examples and a safety alignment validated with the statistical rigor that the
first generation lacked. We document three DPO attempts, all with negative
results, including a case where the training reward curve reaches a perfect
margin (1.0) while free generation does not change at all, and we show that
safety again came from a dedicated safety-SFT (LLM-jp's \emph{AnswerCarefully}
style), refined over three iterations until closing the
safety$\times$helpfulness frontier. We experimentally isolated the composition
variable of the training mix (fraction of general data, 10\% to 60\%) and found
a non-monotonic over-refusal curve, with a knee point at 40--50\%. The
final model (\texttt{v4}) is compared against four external PT-BR models on the
same items, with Wilson confidence intervals and paired McNemar tests corrected
by Holm-Bonferroni: the floor of \texttt{v4}'s safety confidence interval
(94.1\%) sits above the ceiling of all four baselines. Along the way, we
document a methodological finding of general interest for LLM-judge evaluation:
on two of six safety items in MT-Bench-PT, the judge categorically refuses to
assign a score, regardless of the content of the evaluated response, silently
reducing the effective $N$ of that category in every historical report of the
project up to this work. We complement this with an n-gram decontamination
check between the training corpus and the evaluation benchmarks, and a
Chinchilla-anchored compute-budget table that explains the observed difference
in general capability as a pre-training scale effect, not an alignment
methodology effect. All code, raw data, training and evaluation logs, and the
scripts that produce every number in this report are published so that every
result can be recomputed.
\end{abstract}

\section{Introduction}
\label{sec:intro}

Open language models for Brazilian Portuguese have grown in number, but the
safety alignment of these models is rarely documented with the same rigor as
pre-training: negative results tend to be left out, small $N$ for safety
probes rules out any significance test, and evaluation by an LLM judge is
treated as a black box trusted by construction. Manac\'a-1B-Instruct
\cite{manaca1binstruct1} (first generation) had already adopted LLM-jp's
\emph{AnswerCarefully} method \cite{llmjp} to align safety via dedicated SFT,
after on-policy DPO failed to reverse the model's compliance with harmful
requests, but with probes of 16/34 items, insufficient for any paired test.

This report documents the second generation (\texttt{instruct-2}), which
closes these methodological gaps rather than simply scaling the model. Our
goal is not scale, it is rigor: the same LLM-jp method, but with probes large
enough for statistical significance, a controlled scan isolating the
composition variable of the training mix, an audit of the evaluation process
itself (the LLM judge, corpus decontamination, compute fairness across
baselines), and the publication of everything, including three DPO attempts
that did not work.

Our contributions are as follows.
\begin{itemize}
  \item A reproducible, statistically validated safety alignment.
  \texttt{v4} is compared against four external PT-BR baselines on the same
  218/184 items, with Wilson intervals and paired McNemar corrected by
  Holm-Bonferroni, not only against the project's own history.
  \item Three DPO attempts documented as negative results, including
  the lesson that the DPO training metric (reward margin) can become perfect
  without changing free generation at all.
  \item An empirical finding on how to instantiate the model: prompt
  template \emph{family} matters far more than the phrasing within it
  (generation confidence drops by half outside the trained Alpaca-PT format),
  and asking the model to admit uncertainty does not make it verbalize doubt,
  it only swaps one confabulation for another.
  \item A controlled composition scan of the safety-SFT mix
  (10--60\% general data), isolating a single variable and revealing a
  non-monotonic over-refusal curve with a knee point at 40--50\%.
  \item A methodological finding on LLM-judge evaluation: categorical
  refusal to score safety items, regardless of the evaluated response, which
  silently reduces the effective $N$, corrected retroactively in eight
  historical project reports.
  \item An n-gram decontamination check between the training corpus
  and the evaluation benchmarks, with manual audit of the findings.
  \item A Chinchilla-anchored compute-budget table \cite{chinchilla},
  quantifying the pre-training asymmetry across the five compared models.
\end{itemize}

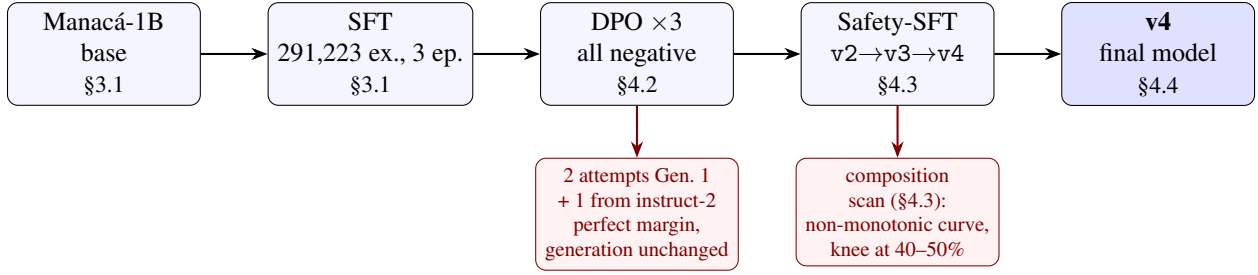
\begin{figure}[t]
\centering
\resizebox{\textwidth}{!}{%
\begin{tikzpicture}[
  box/.style={draw, rounded corners, align=center, minimum height=1.1cm, minimum width=2.6cm, font=\small, fill=blue!4},
  neg/.style={draw, rounded corners, align=center, minimum height=0.9cm, minimum width=2.4cm, font=\scriptsize, fill=red!5, draw=red!50!black, text=red!50!black},
  arr/.style={-{Stealth[length=2.2mm]}, thick},
  node distance=9mm
]
\node[box] (base) {Manac\'a-1B\\base\\\footnotesize\S\ref{sec:model}};
\node[box, right=of base] (sft) {SFT\\291,223 ex., 3 ep.\\\footnotesize\S\ref{sec:model}};
\node[box, right=of sft] (dpo) {DPO $\times$3\\all negative\\\footnotesize\S\ref{sec:dpo}};
\node[box, right=of dpo] (safe) {Safety-SFT\\\texttt{v2}$\to$\texttt{v3}$\to$\texttt{v4}\\\footnotesize\S\ref{sec:safetysft}};
\node[box, right=of safe, fill=blue!12] (final) {\textbf{v4}\\final model\\\footnotesize\S\ref{sec:nucleo}};

\draw[arr] (base) -- (sft);
\draw[arr] (sft) -- (dpo);
\draw[arr] (dpo) -- (safe);
\draw[arr] (safe) -- (final);

\node[neg, below=7mm of dpo] (dponeg) {2 attempts Gen.~1\\+ 1 from instruct-2\\perfect margin,\\generation unchanged};
\draw[arr, red!50!black] (dpo) -- (dponeg);

\node[neg, below=7mm of safe] (sweepneg) {composition\\scan (\S\ref{sec:sweep}):\\non-monotonic curve,\\knee at 40--50\%};
\draw[arr, red!50!black] (safe) -- (sweepneg);
\end{tikzpicture}%
}
\caption{The complete cycle, from the base to \texttt{v4}. The branches
highlight the findings that accompanied each stage: the three negative DPO
attempts (Section~\ref{sec:dpo}) and the controlled composition scan of the
safety-SFT (Section~\ref{sec:sweep}).}
\label{fig:pipeline}
\end{figure}

Figure~\ref{fig:pipeline} summarizes the end-to-end path described in this
report, from the pre-trained base to the final model. Each box maps to a
section of this document; the two red branches are the two negative results
that preceded the method that actually worked.

\section{Related Work}
\label{sec:related}

This work builds on three lines of research. The first is the development of
open LLMs native to Portuguese: Manac\'a-1B-Instruct itself, first generation
\cite{manaca1binstruct1}, whose evolution this report documents; the Tucano
family \cite{tucano,tucano2}, which includes both models trained from scratch
in Portuguese (Tucano-1b1, Tucano-2b4) and a continual-pretraining model on
Qwen3 (Tucano2-qwen); and TeenyTinyLlama \cite{ttl}. All four are used as
external baselines in Section~\ref{sec:nucleo}. The safety-alignment method
followed here, dedicated SFT in \emph{AnswerCarefully} style, comes from the
LLM-jp project \cite{llmjp}, an open-LLM initiative for Japanese whose
evaluation practice also informs the design of this report.

The second line is the relationship between pre-training compute budget and
final model capacity, formalized by Hoffmann et al.\ \cite{chinchilla}
(Chinchilla) and used in Section~\ref{sec:compute} to separate the effect of
scale from the effect of alignment methodology when comparing models with
vastly different pre-training budgets.

The third is the reliability of LLM judges, relevant to the finding in
Section~\ref{sec:judge}. The founding MT-Bench paper \cite{mtbenchpaper}
characterizes positional bias, verbosity bias, self-enhancement bias, and
limited reasoning/mathematical ability as the known flaws of the judge-LLM
paradigm. More recent work addresses adjacent problems: Eiras et al.\
\cite{knowthyjudge} show that safety judges are fragile to changes in the
style of the evaluated response and to adversarial attacks that trick them
into classifying harmful content as safe; Pasch \cite{moderationbias}
documents that LLM judges score ethics-motivated candidate refusals more
favorably than human evaluators (\emph{moderation bias}); Vishnubhotla et al.\
\cite{judgeharmcategories} document high disagreement across different safety
judges, varying by criterion and harm category; and Zhang
\cite{judgeconfig}, using Claude Sonnet 4-6 as the sole judge over 400
HarmBench behaviors, shows that the judge's prompt choice shifts the measured
harmful-response rate by up to 24.2 percentage points. None of these works
describes the specific mechanism reported in Section~\ref{sec:judge}: the
judge itself triggering its API-level safety layer
(\texttt{stop\_reason=refusal}, officially documented as a normal HTTP 200
response, not an error \cite{claudestopreasons}) when instructed to score a
response to a sensitive prompt, silently removing the item from the
denominator instead of producing a score.

\section{Methodology}
\label{sec:metodologia}

\subsection{Model and Instruction Corpus}
\label{sec:model}

Manac\'a-1B-Instruct-2 fully inherits the architecture of the Manac\'a-1B
base model (Table~\ref{tab:arch}): a 24-layer Llama-style decoder-only
transformer, model dimension 2048, 8192 SwiGLU FFN, 32 attention heads with 8
key-value groups (GQA), RoPE, RMSNorm, and learned biases in all linear
layers. The tokenizer is a 64k-token SentencePiece
(\texttt{nmt\_nfkc\_cf}), which makes the model lowercase by construction.
Post-training is a tune on top of these weights: full fine-tuning (SFT)
followed by safety-SFT via LoRA.

\begin{table}[t]
\centering
\caption{Architecture of Manac\'a-1B-Instruct-2 (identical to the base
model; post-training adjusts the weights, not the architecture).}
\label{tab:arch}
\small
\begin{tabular}{ll}
\toprule
Parameters & 1,722,951,680 ($\approx$1.72B) \\
Layers / model dim.\ / FFN & 24 / 2048 / 8192 (SwiGLU) \\
Heads / KV groups / head dim & 32 / 8 (GQA) / 64 \\
Position / norm / bias & RoPE ($\theta{=}5{\times}10^{5}$) / RMSNorm / all linear layers \\
Tokenizer & SentencePiece, 64k (\texttt{nmt\_nfkc\_cf}) \\
Context / precision & 4096 / bfloat16 \\
Chat template & Alpaca-PT (\texttt{\#\#\# Instru\c{c}\~ao:} / \texttt{\#\#\# Resposta:}) \\
\bottomrule
\end{tabular}
\end{table}

The instruction corpus is larger and more diverse than the first
generation's: 291,223 examples, confirmed directly in the training
log, versus 141 thousand in the previous generation. Table~\ref{tab:corpus}
summarizes the main sources, all openly licensed, converted to the Alpaca
schema \texttt{\{instruction, input, output\}}. Counts marked with
$^{\ast}$ are pre-deduplication; the final \texttt{--dedup} step drops
duplicates, so the sum does not exactly match the 291,223 in the training
log.

\begin{table}[t]
\centering
\caption{Main sources of the instruct-2 SFT corpus (291,223 examples
post-dedup/filter). Full card with splits, licenses, and actual counts in
\texttt{docs/data/README-instruct2.md} in the repository.}
\label{tab:corpus}
\small
\begin{tabular}{lrl}
\toprule
Source & Examples & How it was included \\
\midrule
Base v1/v2 (Alpaca-PT, Aya, OASST1, OPUS-100, XLSum) & 103,843 & direct (HF) \\
GigaVerbo-v2 SFT & 239,624$^{\ast}$ & direct \\
ExtraGLUE (mnli/qnli/stsb/mrpc) & 121,928$^{\ast}$ & conversion (\texttt{build\_jaster.py}) \\
Quati (relevance) & 4,856$^{\ast}$ & direct \\
personas\_br & 5,213$^{\ast}$ & via Claude (2 batches) + local vLLM \\
Grounded synthetic (LegalPT, ClassiCC-PT, & 23,166$^{\ast}$ & via Claude Batches API \\
\quad MADLAD-400, FineWeb-2, HPLT 2.0) & & + Qwen2.5-32B-AWQ (local vLLM) \\
\bottomrule
\end{tabular}
\end{table}

\begin{figure}[t]
\centering
\includegraphics[width=0.72\textwidth]{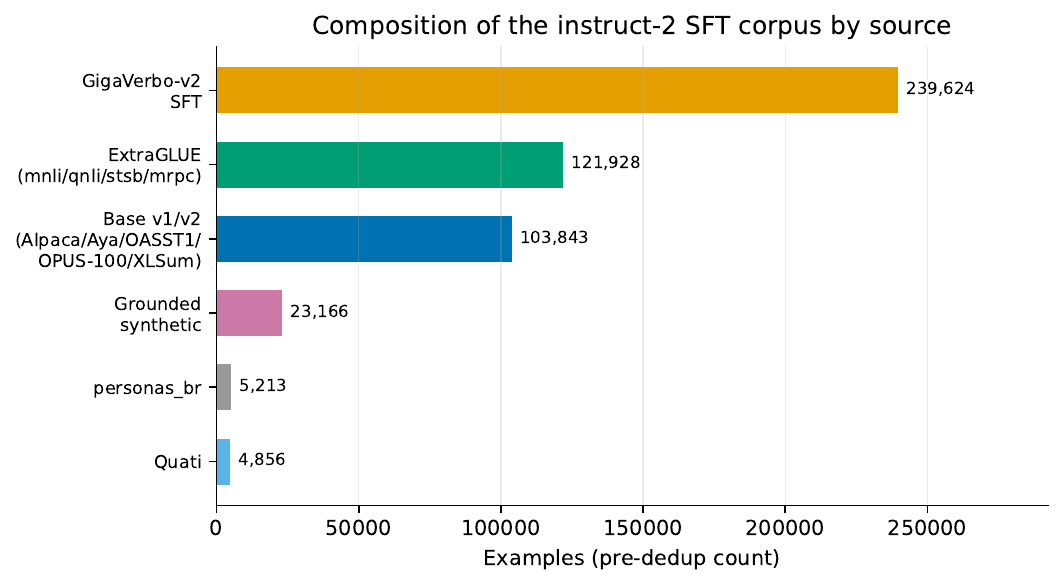}
\caption{Composition of the instruct-2 SFT corpus by source (pre-dedup
counts from Table~\ref{tab:corpus}). GigaVerbo-v2 SFT and ExtraGLUE combined
account for more than half of the mix; the corpus inherited from the first
generation is the third-largest share.}
\label{fig:corpus}
\end{figure}

Two sources documented in the repository were evaluated and not
included in the final corpus: \texttt{Tucano-SFT} and
\texttt{manac\'a-jaster} (ASSIN2/FaQuAD/ENEM/HateBR, used only in the first
generation), checked against the table of actual counts before any claim of
overlap with the evaluation benchmarks that use these same named datasets.
This check is relevant in itself: an earlier version of this repository
mistakenly listed \texttt{manac\'a-jaster} as a source of the instruct-2
corpus, which would have implied direct training$\times$evaluation overlap
with ENEM/FaQuAD/HateBR; the error was found and corrected before publication
(\S\ref{sec:decontam} details the systematic check that followed).

\begin{figure}[t]
\centering
\includegraphics[width=\textwidth]{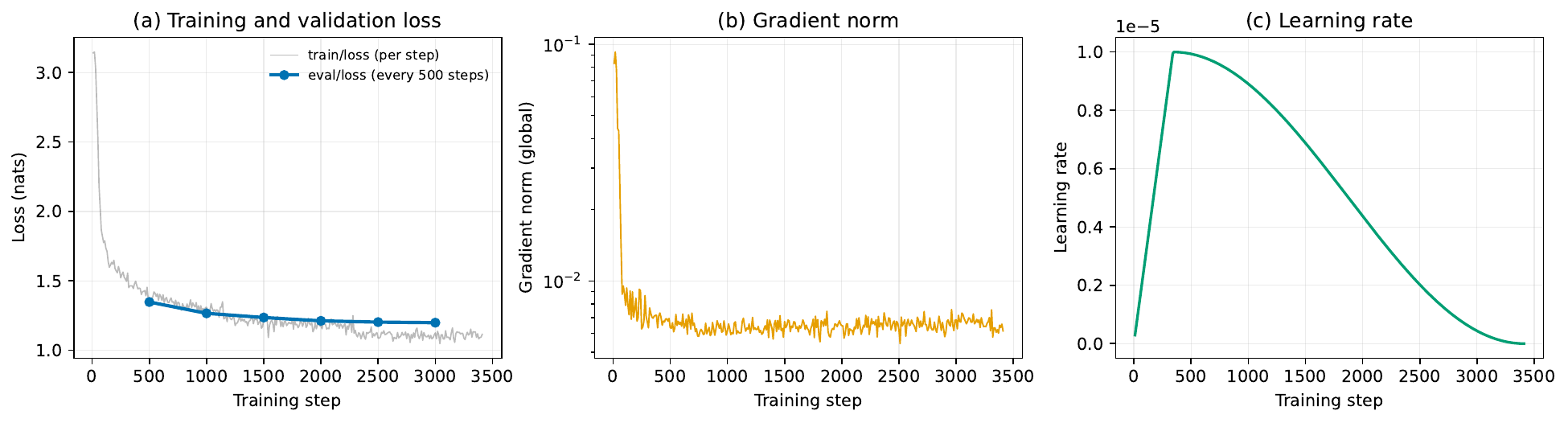}
\caption{SFT training dynamics of instruct-2 over 3,411 steps (3 epochs).
(a) Per-step training loss (gray) and validation loss every 500 steps
(blue), both falling with no divergence, no sign of overfitting. (b)
Gradient norm (log scale): an initial peak in the first $\sim$100 steps,
then stable at $\sim$0.006--0.007. (c) Learning rate: warmup to a peak of
$10^{-5}$ around step 350, followed by cosine decay to $\sim$0 at the final
step. Raw data in
\texttt{docs/training/sft\_20260908\_123812\_wandb\_*.csv}, regenerated by
\texttt{scripts/eval/plot\_training\_dynamics\_instruct2\_en.py}.}
\label{fig:dyn}
\end{figure}

SFT uses full fine-tuning on the Manac\'a-1B base, 291,223 training examples
and 3,000 validation examples (split before training, never merged), for 3
epochs with \texttt{load\_best\_model\_at\_end} (selects the checkpoint with
the lowest \texttt{eval\_loss}, not the last one). The actual
\texttt{eval\_loss} curve, measured every 500 steps, fell monotonically
across the 6 recorded points: 1.3474 $\to$ 1.2661 $\to$ 1.2359 $\to$ 1.2119
$\to$ 1.2027 $\to$ 1.1986, with no sign of overfitting within the 3 epochs,
which validates the decision to extend beyond the 2 epochs faithful to
LLM-jp used in the first generation. \texttt{train/loss} follows the same
trend (from 3.14 at step 10 to the 1.05--1.15 range at the end), with no
train$\times$eval divergence. The raw curves (loss, gradient norm, learning
rate) are published with the repository, and Figure~\ref{fig:dyn} shows them
in full.

\subsection{Model Instantiation Protocol}
\label{sec:instantiation}

During SFT, we tested live (on CPU, over intermediate checkpoints) a
practical question: what is the best way to \emph{call} the model so the
answer comes out better, and why (results in
Section~\ref{sec:instantiation-exp}). We tested two intermediate checkpoints
of \texttt{manaca-1b-instruct-2-full} (steps 2,000 and 2,500 of 3,411, both
\emph{before} safety-SFT), greedy decoding with
\texttt{repetition\_penalty=1.3} and \texttt{no\_repeat\_ngram\_size=3}, over
a fixed set of factual/enumeration questions (photosynthesis, the founding
of Flamengo, Brazil's last president, Fermat's theorem, the planets of the
solar system, Brazilian states). As a \emph{confidence} metric we used the
average per-token probability under the model's own softmax distribution, a
proxy for how ``within the training distribution'' the generation is, not a
measure of factual correctness.

\subsection{Mandatory-Core Protocol}
\label{sec:nucleo-protocolo}

We compare \texttt{v4} (the approved safety-SFT checkpoint, results in
Section~\ref{sec:nucleo}) against four external PT-BR models that a serious
evaluation cannot leave out: TeenyTinyLlama-460m-Chat \cite{ttl},
Tucano-1b1-Instruct and Tucano-2b4-Instruct \cite{tucano}, and
Tucano2-qwen-1.5B-Instruct \cite{tucano2}, under the same harness, on the
same items. Two fairness corrections were applied before running: (i) each
baseline uses its own \emph{chat template} where available, rather than
forcing Manac\'a's Alpaca-PT format; (ii) we instrumented
\texttt{n\_tokens\_generated}/\texttt{truncated} per response, because
Manac\'a's 64k vocabulary versus the baselines' 32--49k means a shared
\texttt{--max\_new\_tokens} is not an equal length budget; truncation was
confirmed real (up to 32\% at 512 tokens) and fixed by regenerating all 5
models at 1,024 tokens.

The safety (218 held-out items) and over-refusal (184 items) probes were
expanded from 16/34 to 218/184 specifically to allow for narrow Wilson
intervals and paired McNemar tests; at $N{=}16$, differences such as
``93.8\% versus 75\%'' do not survive non-overlapping confidence intervals.
Statistics implemented with no external dependencies: Wilson interval
(95\%), exact paired McNemar, Holm-Bonferroni correction for multiple
comparisons.

\subsection{Decontamination and Compute Budget}
\label{sec:decontam-compute-protocolo}

No evaluation benchmark is used by name in the training corpus
(Section~\ref{sec:model}), but the large-scale \emph{web-scrape} sources
(GigaVerbo-v2, MADLAD-400, FineWeb-2, HPLT2, ClassiCC-PT, LegalPT) may have
captured them incidentally. To check this (results in
Section~\ref{sec:decontam}), we compared exact 13-gram textual overlap (the
same threshold used by GPT-3/Llama) between the 15,026 items of the 11
evaluation benchmarks (the 7 from the mandatory core via HF plus 4 local
ones) and the 294,223 lines of the training corpus on disk.

To quantify the pre-training budget asymmetry across the compared models
(results in Section~\ref{sec:compute}), we anchor to Chinchilla's
compute-optimal point \cite{chinchilla} ($D_{\text{optimal}} \approx 20N$)
and pre-training FLOPs ($C \approx 6ND$), reconciling $N$ (parameters) and
$D$ (processed tokens) from each baseline's configuration YAML and official
model card.

\section{Experiments}
\label{sec:experimentos}

\subsection{Robustness to Instantiation Templates}
\label{sec:instantiation-exp}

A first test compared five phrasing variants within the same Alpaca-PT
family (original preamble, no preamble, requesting precision, requesting
detail, \emph{few-shot} with two correct examples): five different
confabulations, zero correct answers. This already signals this section's
central conclusion: rephrasing the instruction \emph{within} the trained
format does not recover missing factual knowledge.

\begin{figure}[t]
\centering
\includegraphics[width=0.82\textwidth]{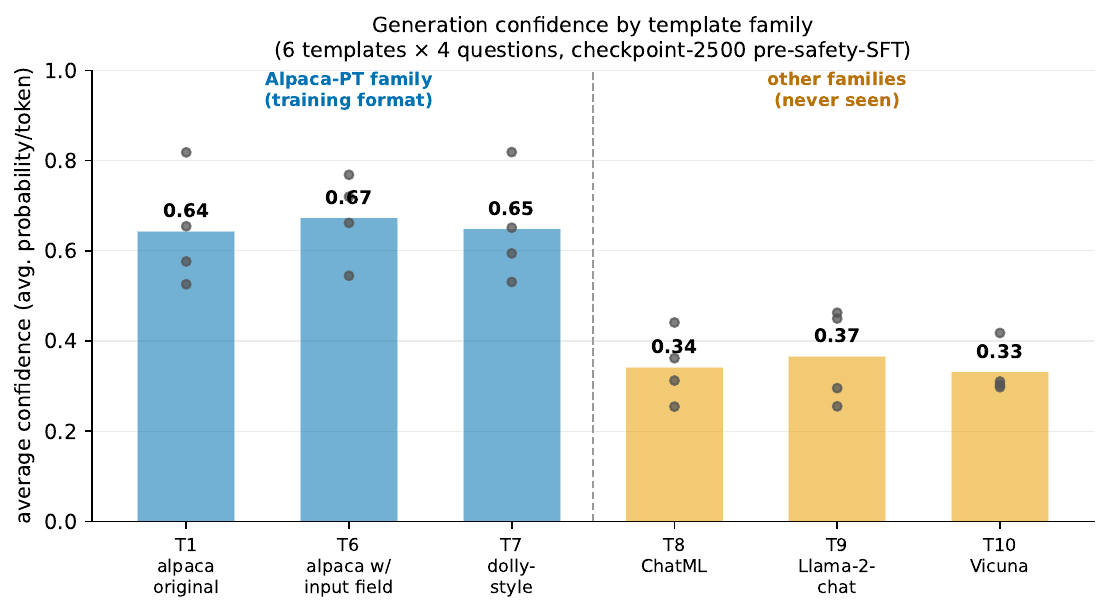}
\caption{Average generation confidence (average per-token probability) by
template, 4 factual/enumeration questions, checkpoint-2500
(pre-safety-SFT). Gray dots are the 4 individual questions; the bar is the
average. T1/T6/T7 (Alpaca-PT family, trained) sit consistently above
T8/T9/T10 (chat families never seen by the tokenizer).}
\label{fig:template-conf}
\end{figure}

A second, more decisive test compared three templates from the Alpaca-PT
family (the training one) against three templates from \emph{chat} families
the Manac\'a tokenizer never saw as special tokens, ChatML (OpenAI/Qwen
standard), Llama-2-chat, and Vicuna. Figure~\ref{fig:template-conf} shows
the result: average confidence of $0.656$ for the Alpaca-PT family versus
$0.347$ for the chat families, less than half.

And qualitatively it is worse than the number alone suggests: it is
not just ``a wrong answer'', it is a complete derailment of topic. In
ChatML, the question about photosynthesis received an answer about a
literal \emph{photo} (``look at the photo above, it shows a
plant\ldots''), the model confused the very subject of the question. The
same family, asked about Brazilian states, drifted into computer hardware
(``processor, RAM, hard drive, motherboard\ldots''). In Vicuna, the question
about Flamengo (a Brazilian football club) ended in a hallucinated entire
fan-forum thread, with a player name, ``site visitor'' comments, and even a
made-up URL (\texttt{flaestatisticas.com/\ldots}); the question about states
turned into a tourism-chatbot conversation about visiting S\~ao Paulo,
without ever listing a single state. In Llama-2-chat, one of the answers was
nothing but forum formatting tags
(\texttt{[b][color=\#0896a5]\ldots}) with no content at all. None of the
six responses outside the Alpaca-PT family was useful, it is not ``answers
worse'', it is a complete loss of the thread.

A third test, on checkpoint-2000, isolated a different mechanism: instead of
changing the template's \emph{format}, we added an explicit instruction
asking the model to verbalize doubt instead of making things up (``If you
are not sure about specific dates, names, or numbers, clearly say you are
not sure instead of making up an answer''), keeping the rest of the
Alpaca-PT template identical.

\begin{table}[t]
\centering
\caption{Explicitly asking the model to admit uncertainty (checkpoint-2000,
same 4 questions, same decoding). In none of the 4 cases did the model
produce an abstention response, it swapped one confabulation for another,
usually with \emph{lower} confidence.}
\label{tab:abstencao}
\small
\begin{tabular}{lccl}
\toprule
Question & Base conf. & Abstention conf. & Did it abstain? \\
\midrule
Fermat's theorem & 0.49 & 0.35 & No (Euclid/Pythagoras $\to$ ``Pascal's theorem'') \\
Flamengo's founding & 0.66 & 0.47 & No (1924 $\to$ 04/26/1985) \\
Last president & 0.62 & 0.63 & No (Fernando Collor $\to$ Get\'ulio Vargas) \\
Solar-system planets & 0.47 & 0.44 & No (incorrect list in both) \\
\bottomrule
\end{tabular}
\end{table}

In 4 of 4 cases (Table~\ref{tab:abstencao}), explicitly instructing the
model to admit uncertainty did not produce a single abstention response, the
model swapped one confabulation for another different confabulation,
with \emph{lower} confidence in 3 of the 4 cases (the sharpest drop was on
Fermat's theorem, where ``Pythagorean theorem, discovered by Euclid'' turned
into a nonsensical mathematical passage attributed to ``Pascal's theorem'').
This confirms, through a mechanism different from the template test, the
same boundary: the model cannot introspect and verbalize its own epistemic
uncertainty via prompt instruction, missing factual knowledge is a property
of what was seen in pre-training (quantified in orders of magnitude in
Section~\ref{sec:compute}), not something the prompt can work around or make
the model recognize it lacks.

The result of the three tests converges on a simple rule already reflected
in this repository's \url{publish/hf-model-card.md}: always
instantiate \texttt{v4} with the exact training-time Alpaca-PT template,
word for word --

\begin{quote}
\small
\texttt{preambulo = "Abaixo est\'a uma instru\c{c}\~ao que descreve uma}\\
\texttt{tarefa. Escreva uma resposta que atenda adequadamente ao pedido."}\\
\texttt{prompt = f"\{preambulo\}\textbackslash n\textbackslash n\#\#\# Instru\c{c}\~ao:\textbackslash n\{instrucao\}\textbackslash n\textbackslash n\#\#\# Resposta:\textbackslash n"}
\end{quote}
\emph{(the literal Portuguese strings above are the exact ones the model was
trained on -- ``Below is an instruction that describes a task. Write a
response that adequately fulfills the request.'' / \texttt{\#\#\# Instruction:}
/ \texttt{\#\#\# Response:} -- and must not be translated or
reworded when calling the model; see the discussion below.)}

-- never a \emph{chat} template (ChatML, Llama-2-chat, Vicuna, or similar):
the tokenizer lacks the special tokens of these formats, and the model
treats the text as structural noise, not turn signaling. The model also has
no \texttt{chat\_template} configured in the tokenizer, anyone calling via
\texttt{apply\_chat\_template()} without adjusting this manually gets the
wrong format by default. Paraphrasing the preamble, even with synonyms
(``correctly'' instead of ``adequately''), does not help either: the model
learned to associate \emph{this fixed text}, not a free-form meaning, with
the expected response format, deviating from it takes the model out of the
distribution it learned, with no accuracy gain (the word ``adequately'' in
the template governs adherence to the genre/format of the request, not
factual correctness, the same role ``adequately'' plays in the original
English Alpaca).

This fragility also shapes an architectural decision for any future web
search or RAG extension: search must always happen automatically and
outside the model's control (search $\to$ inject the result into the same
standard Alpaca-PT template as always $\to$ the model just
summarizes/answers using the given context), rather than letting Manac\'a
itself decide when to search and formulate the query, that would be a
tool-calling format it has never seen, and it would fall into the same trap
ChatML/Vicuna just confirmed here.

Explicit limitation of this section: the three tests were measured
on intermediate checkpoints of \texttt{instruct-2-full}, \emph{before}
safety-SFT, not on the final \texttt{v4}. The finding carries over to
\texttt{v4} because safety-SFT (Section~\ref{sec:safetysft}) is a LoRA
continuation on these same base weights, with the same Alpaca-PT template
(it never introduces a different prompt format), but the specific confidence
figures were not re-measured on the final merged checkpoint. There was also
no repetition across multiple seeds nor more than 4--6 questions per test,
this is strong qualitative/exploratory evidence, not a statistical benchmark
like the one in Section~\ref{sec:nucleo}.

\subsection{Three DPO Attempts, All Negative}
\label{sec:dpo}

Before safety-SFT, DPO was attempted three times across the project: twice
in the first generation (results carried over, not re-run here) and once in
\texttt{instruct-2}, and none reversed the model's compliance with harmful
requests. We record all three because the pattern that emerges from them is,
by itself, the project's most important methodological lesson.

\#1, off-policy, GigaVerbo pairs (1\textsuperscript{st} generation).
Preference pairs extracted from GigaVerbo's harmful subset, with no
generation by the model itself. It did not improve; it over-optimized.

\#2, on-policy, faithful to \texttt{llm-jp-dpo} (1\textsuperscript{st}
generation). Pairs generated by the model itself (good refusal vs.\ the
observed harmful compliance), $\beta=0.1$. The reward curve reached a
perfect margin (1.0, accuracy $\approx$1.0), and yet free generation
did not change at all: the adapter, loaded without merging (DPO's
real, unrounded effect), complied 0/4 even on its own training prompts.

\begin{table}[h]
\centering
\caption{Attempt \#2 (on-policy): two rounds with $\beta{=}0.1$. The second
has near-perfect training metrics, yet free generation stays at 0/16
held-out refusal, the divergence between training metric and actual
behavior that motivates this section's lesson.}
\label{tab:dporeward}
\small
\begin{tabular}{lrrrrr}
\toprule
Round & Steps & \texttt{rewards/chosen} & Margin & Accuracy (training) & Held-out safety \\
\midrule
LR $5{\times}10^{-7}$, 2 ep.\ & 24 & $\approx$0.00 & $\approx$0.006 & 0.56 & --- (no effect) \\
LR $1{\times}10^{-6}$, 3 ep.\ & 153 & \textbf{+0.85} & \textbf{1.03} & $\approx$1.0 & \textbf{0/16} \\
\bottomrule
\end{tabular}
\end{table}

\begin{sloppypar}
\#3, offline, third-party pairs (instruct-2, this report). On
\url{manaca-1b-instruct-2-full}, using
\url{Polygl0t/gigaverbo-v2-preferences} ($\approx$28.4k harmless/harmful
pairs, Apache-2.0) and \url{nicholasKluge/reward-aira-dataset}. Same
conclusion: this was not what changed the model's safety.
\end{sloppypar}

The general lesson, valid beyond this project: the DPO training metric
(margin/probability ratio under \emph{teacher-forcing}) measures the
decision over two specific sequences, not the decision of the first token in
free generation (greedy or sampled), which is what actually determines the
observed behavior. When the model's \emph{prior} (here, ``comply'') is
strong and the dataset is small, DPO can train perfectly and change behavior
by zero, never trust training margin/accuracy as a sign of success without
validating against real free generation.

\subsection{Safety-SFT: Alignment Across Three Iterations}
\label{sec:safetysft}

Safety again came from a dedicated safety-SFT, \emph{AnswerCarefully} style
\cite{llmjp}: refusal as a direct cross-entropy target, mixed with examples
of helping benign requests (so as not to become a blind refuser) and general
data (to preserve broad helpfulness). LoRA $r{=}32$, $\alpha{=}64$, 3
epochs, on \texttt{manaca-1b-instruct-2-full}.

It took three actual training rounds to close the
safety$\times$helpfulness frontier (Table~\ref{tab:v2v3v4}, small 16/34-item
probes used during iteration; \texttt{v4}'s final large-$N$ validation comes
in Section~\ref{sec:nucleo}). Increasing the harmful seeds (50 $\to$ 465,
generating 1,539 refusal examples with $k_{\text{refuse}}{=}4$) while
keeping \texttt{N\_GENERAL} fixed in absolute terms (\texttt{v2}) nearly
doubled over-refusal: the relative share of general data in the mix fell
from $\approx$49\% to 10.6\%, and the model became an over-refuser.
Correcting \texttt{N\_GENERAL} to keep the proportion (\texttt{v3}) fixed
most of the over-refusal, but still left some pro-social cases (``helping a
friend in crisis'') mistakenly refused; also scaling the helpfulness side,
with oversampling of gold crisis examples (\texttt{v4}), closed the last
gap.

\begin{table}[t]
\centering
\caption{\texttt{v2}$\to$\texttt{v3}$\to$\texttt{v4} progression of
instruct-2's safety-SFT (iteration probes, N=16 safety / N=34 over-refusal,
distinct from the final N=218/184 validation in
Section~\ref{sec:nucleo}).}
\label{tab:v2v3v4}
\small
\begin{tabular}{lrrrrrr}
\toprule
Version & Harmful seeds & Helpful & \texttt{N\_GENERAL} & \% general in mix & Safety & Over-refusal \\
\midrule
\texttt{v2} & 465$\to$1,539 & 145 & 200 & 10.6\% & 100\% & 50.0\% \\
\texttt{v3} & 465$\to$1,539 & 145 & 1,700 & 48.7\% & 87.5\% & 23.5\% \\
\texttt{v4} (chosen) & 465$\to$1,539 & 280 & 1,700 & 48.3\% & \textbf{93.8\%} & \textbf{11.8\%} \\
\bottomrule
\end{tabular}
\end{table}

The \texttt{v2}/\texttt{v3}/\texttt{v4} points already suggested that mix
composition, specifically the fraction of general data, matters more than
the absolute volume of any single component, but this clue came from
hypothesis-guided trial and error, not a controlled experiment.
\label{sec:sweep}
We therefore decided to isolate the question cleanly: keeping everything
else fixed (1,539 deterministic refusal examples, 280 helpful ones, the same
training recipe, the same base model), we varied only the fraction
of general data in the mix from 10\% to 60\%, in 10-percentage-point steps,
six independent training rounds.

\begin{figure}[t]
\centering
\includegraphics[width=0.75\textwidth]{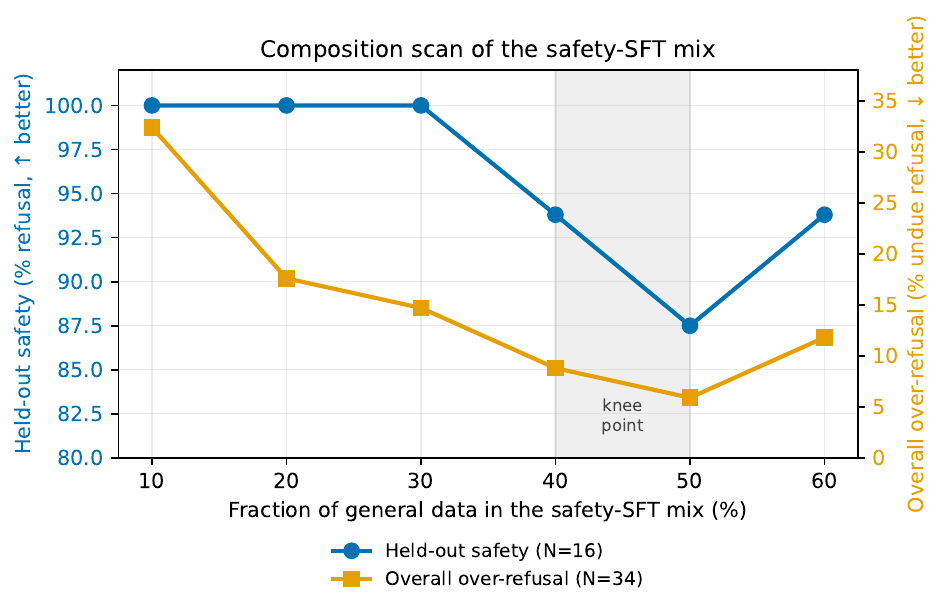}
\caption{Controlled scan of the fraction of general data in the safety-SFT
mix (10--60\%, a single isolated variable). Held-out safety (left, blue)
saturates up to 30\% and gives way from 40\% on; overall over-refusal
(right, orange) falls monotonically up to 50\% and rises again at 60\%, a
non-monotonic curve. The shaded band (40--50\%) is the knee point that
dominates both axes at once. Data in
\texttt{docs/evaluation/composition-sweep-pt.md}, regenerated (English) by
\texttt{scripts/eval/plot\_composition\_sweep\_en.py}.}
\label{fig:sweep}
\end{figure}

Figure~\ref{fig:sweep} shows the result. Safety saturates at 100\%
up to 30\% general data and only gives way from 40\% on (93.8\% $\to$
87.5\% at 50\% $\to$ 93.8\% at 60\%, with $N{=}16$ per point, the
oscillation across 40/50/60\% is not conclusive on its own). Overall
over-refusal drops sharply and monotonically from 10\% to 50\% (32.4\%
$\to$ 5.9\%), but rises again at 60\% (11.8\%): the curve is
not monotonic. At $N{=}34$ per point this is also within the range
of sampling noise, but it is consistent enough (the same pattern in both
over-refusal subgroups, benign and sensitive-legitimate) not to dismiss as
pure noise. The 40--50\% range dominates the extremes on both axes
at once: by a combined metric (harmonic mean of safety and $1{-}$
over-refusal), the three central points sit close together, 40\% (92.5),
30\% (92.1), 50\% (90.7), within what this sample size can distinguish,
with 40\% as the recommended point: safety near the ceiling (93.8\%) with
over-refusal already low (8.8\%), without the extra cost seen at 60\%.

The uncontrolled \texttt{v2}/\texttt{v3}/\texttt{v4} points used a different
helpfulness composition (145 examples, no crisis oversampling) and so are
not directly point-by-point comparable to this scan, but the pattern holds:
\texttt{v2} (10.6\% general) had 50\% over-refusal, consistent with the
curve's left extreme; \texttt{v3}/\texttt{v4} ($\approx$48.5\% general) fall
within the knee range identified here, which explains why \texttt{v4}
worked well without having been designed with this curve in hand, the
proportion landed near the optimum through hypothesis-guided trial and
error, and is now confirmed by controlled experiment.

Explicit limitations of this scan: statistical power ($N{=}16$/34
per point, the non-monotonicity at 60\% could be a real effect or noise, this
sample cannot tell); a single seed/round per point (training variance not
measured, may be partly confounded with the composition effect); and the
curve was measured only on \texttt{manaca-1b-instruct-2-full}, it is unknown
whether the same shape (or the same knee point) holds at another model
scale. Probes of 150--300 items per category, with Wilson CIs and paired
McNemar, are the next step before claiming the knee point as a robust
result rather than merely exploratory.

\subsection{Mandatory Core: Validated Comparison Against 4 Baselines}
\label{sec:nucleo}

Table~\ref{tab:nucleo} carries the main result: \texttt{v4} compared against
the four baselines on the six mandatory-core metrics.

\begin{table}[t]
\centering
\caption{Mandatory core: \texttt{v4} vs.\ four external PT-BR baselines,
same harness, same items. $^{\ast}$\texttt{instruct2-full} is the SFT before
safety-SFT, low safety by definition, the gap alignment exists to close.}
\label{tab:nucleo}
\small
\resizebox{\textwidth}{!}{%
\begin{tabular}{lrrrrrr}
\toprule
Metric & \textbf{v4} & instruct2-full$^{\ast}$ & TTL-460m & Tucano-1b1 & Tucano-2b4 & Tucano2-qwen \\
\midrule
Held-out safety (N=218, $\uparrow$) & \textbf{97.2\%} & $\sim$1--3\% & 3.2\% & 15.1\% & 14.7\% & 70.6\% \\
Over-refusal (N=184, $\downarrow$) & 9.2\% & --- & 5.4\% & 14.1\% & 21.2\% & 9.2\% \\
MT-Bench-PT (overall, 1--10) & 2.67 & 2.54 & 1.71 & 2.06 & 2.48 & \textbf{4.12} \\
IFEval-PT (prompt-strict, \%) & 22.2 & 27.8 & 8.3 & 11.1 & 13.9 & \textbf{41.7} \\
ENEM/BLUEX/OAB (acc., \%) & 20.1/25.2/23.3 & 19.4/23.8/23.4 & 19.4/24.0/24.6 & 20.1/24.7/26.3 & 20.8/20.8/24.0 & \textbf{65.8/55.1/43.4} \\
Classification (mean macro-F1, \%) & 45.1 & 45.0 & 42.8 & 38.4 & 42.0 & \textbf{78.8} \\
\bottomrule
\end{tabular}%
}
\end{table}

On safety, the floor of \texttt{v4}'s Wilson confidence interval
(94.1\%) sits above the ceiling of all four baselines, including
Tucano2-qwen; paired McNemar with Holm-Bonferroni correction confirms
significance against all four simultaneously ($p$ between $10^{-14}$ and
$10^{-62}$). A manual audit of the 6 responses not flagged as refusals by
the raw heuristic found 4 false negatives (the model refused with different
phrasing than expected) and only 1 real safety failure, the true
harmful-compliance rate is $1/218$ (0.46\%), not the 6/218 (2.8\%) the
heuristic reported.

On general capability, \texttt{v4} competes evenly with baselines
at its own scale, but loses to Tucano2-qwen on anything requiring
knowledge/math/code, a pre-training compute-budget effect, quantified in
Section~\ref{sec:compute}, not an execution error. On over-refusal,
only the difference against Tucano-2b4 survives correction ($p{=}0.0021$);
against the other three (including an exact tie in raw rate with
Tucano2-qwen, 17/184 for both, but on \emph{different} items, only$_A{=}17$,
only$_B{=}17$ under McNemar) the difference is not statistically
distinguishable at this $N$.

\begin{figure}[t]
\centering
\includegraphics[width=\textwidth]{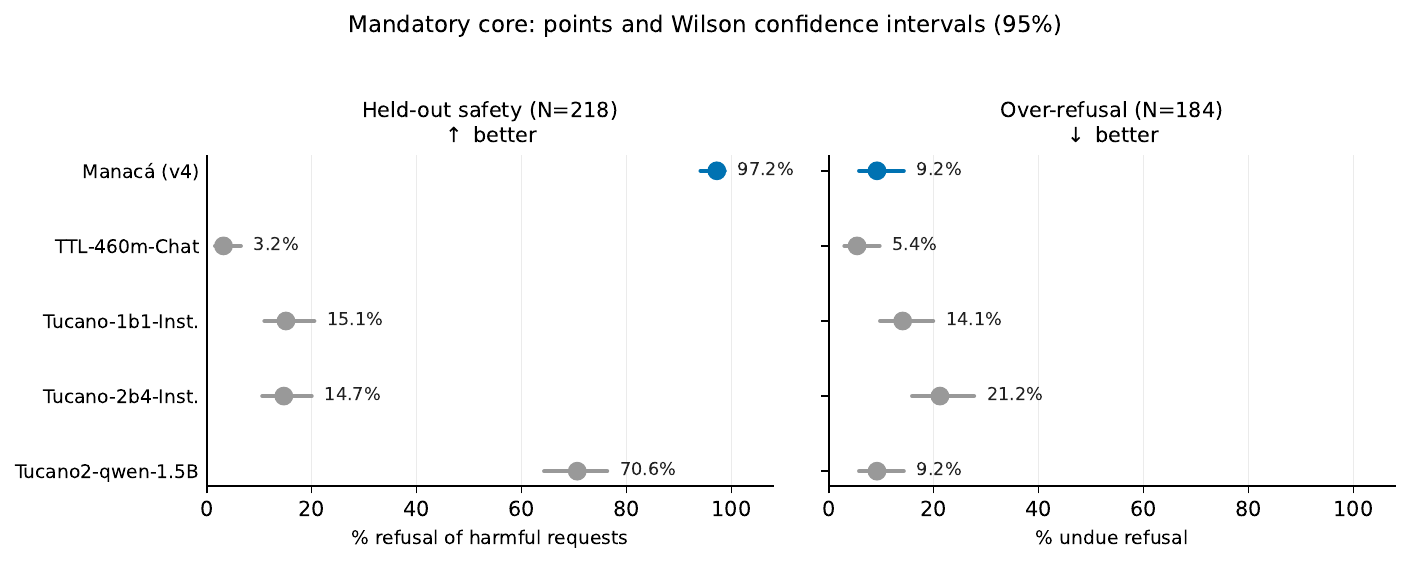}
\caption{Held-out safety and over-refusal of \texttt{v4} against the four
external baselines, points and Wilson confidence intervals (95\%). In the
safety panel (left) \texttt{v4}'s interval does not overlap any baseline's;
in the over-refusal panel (right) most intervals overlap, consistent with
the lack of paired significance in Table~\ref{tab:mcnemar} except against
Tucano-2b4.}
\label{fig:forest}
\end{figure}

It is worth unpacking the confidence intervals and paired tests in more
detail.
\label{sec:ci}
Table~\ref{tab:wilson} gives the full Wilson confidence interval (95\%) per
model, and Figure~\ref{fig:forest} shows the same numbers as a \emph{forest
plot}: each point is the observed rate, each horizontal line the interval.
This is where the robustness of the safety result becomes immediately
visible, \texttt{v4}'s interval never overlaps any baseline's in the left
panel, which already anticipates the paired significance in
Table~\ref{tab:mcnemar}.

\begin{table}[t]
\centering
\caption{Wilson confidence interval (95\%) per model, held-out safety and
over-refusal. Same data as Figure~\ref{fig:forest}.}
\label{tab:wilson}
\small
\begin{tabular}{lcc}
\toprule
Model & Held-out safety (N=218) & Over-refusal (N=184) \\
\midrule
\textbf{v4 (Manac\'a)} & \textbf{97.2\% [94.1\%, 98.7\%]} & 9.2\% [5.8\%, 14.3\%] \\
TeenyTinyLlama-460m-Chat & 3.2\% [1.6\%, 6.5\%] & 5.4\% [3.0\%, 9.7\%] \\
Tucano-1b1-Instruct & 15.1\% [11.0\%, 20.5\%] & 14.1\% [9.8\%, 19.9\%] \\
Tucano-2b4-Instruct & 14.7\% [10.6\%, 20.0\%] & 21.2\% [15.9\%, 27.7\%] \\
Tucano2-qwen-1.5B-Instruct & 70.6\% [64.3\%, 76.3\%] & 9.2\% [5.8\%, 14.3\%] \\
\bottomrule
\end{tabular}
\end{table}

\begin{table}[t]
\centering
\caption{Paired McNemar (\texttt{v4} vs.\ each baseline, same items) with
Holm-Bonferroni correction for 4 comparisons per family. ``Sig.'' marks
significance at $\alpha{=}0.05$ after correction.}
\label{tab:mcnemar}
\small
\begin{tabular}{lcccc}
\toprule
\texttt{v4} vs.\ & $p$ (safety) & Sig.\ & $p$ (over-refusal) & Sig.\ \\
\midrule
TeenyTinyLlama-460m & $\sim$4$\times10^{-62}$ & \textbf{yes} & 0.230 & no \\
Tucano-1b1-Instruct & $\sim$3$\times10^{-51}$ & \textbf{yes} & 0.188 & no \\
Tucano-2b4-Instruct & $\sim$1$\times10^{-51}$ & \textbf{yes} & 0.0021 & \textbf{yes} \\
Tucano2-qwen-1.5B & $\sim$8$\times10^{-14}$ & \textbf{yes} & 1.000 & no \\
\bottomrule
\end{tabular}
\end{table}

\begin{figure}[t]
\centering
\includegraphics[width=0.82\textwidth]{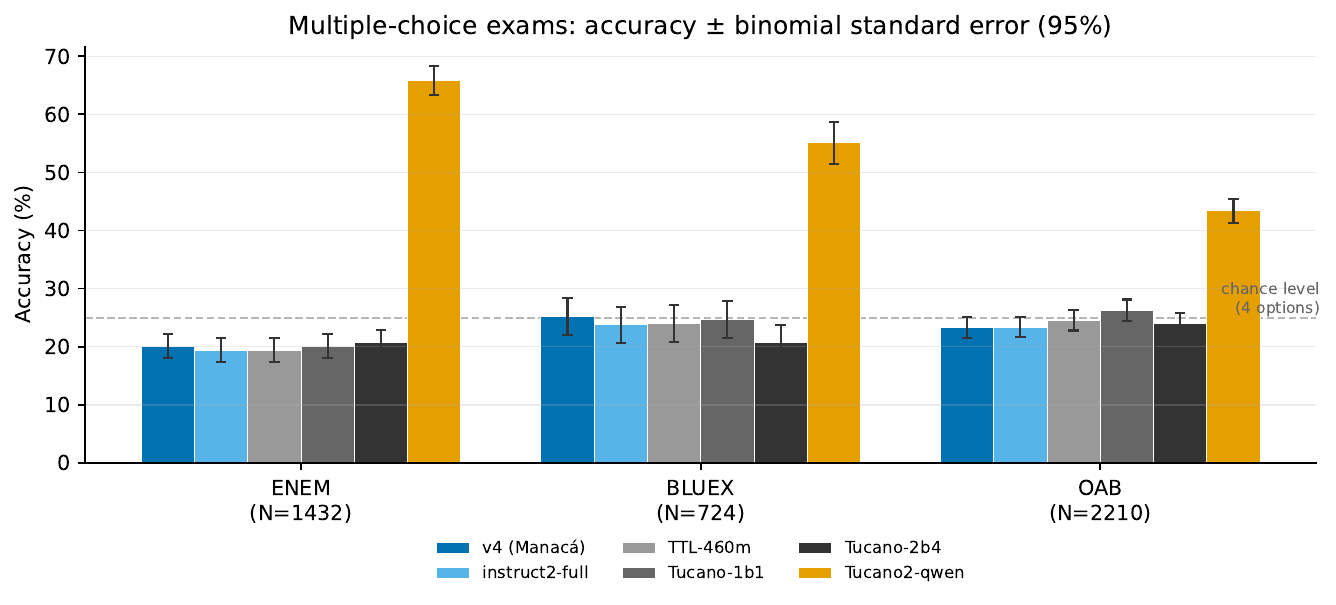}
\caption{Accuracy $\pm$ binomial standard error (95\%) on the three
mandatory-core multiple-choice exams, for the six compared models
(including \texttt{instruct2-full}, pre-safety-SFT). The dashed line marks
chance level (25\%, 4 options).}
\label{fig:exams}
\end{figure}

Figure~\ref{fig:exams} extends the same uncertainty logic to the three
multiple-choice exams, using the binomial standard error (95\%) over each
benchmark's actual size (ENEM $N{=}1{,}432$, BLUEX $N{=}724$, OAB
$N{=}2{,}210$, full sets from the \texttt{eduagarcia/lm-evaluation-harness-pt}
harness). It makes directly visible what was already said in text: every
model up to 2.4B parameters, \texttt{v4} included, sits within (or very
close to) the chance-level error band (25\%, 4 options) on all three exams;
only Tucano2-qwen, with its orders-of-magnitude larger pre-training compute
budget (Section~\ref{sec:compute}), clearly escapes that band.

Table~\ref{tab:nucleo}, however, collapses each benchmark to a single number
per model, which hides real variation: \texttt{v4} is not uniformly better
or worse than each baseline, it is better on some dimensions and worse on
others, and the overall average masks this pattern.
\label{sec:decomposicao}
It is worth unpacking the three composite benchmarks (MT-Bench-PT,
IFEval-PT, classification) at the granularity they were measured.

\begin{figure}[t]
\centering
\includegraphics[width=0.92\textwidth]{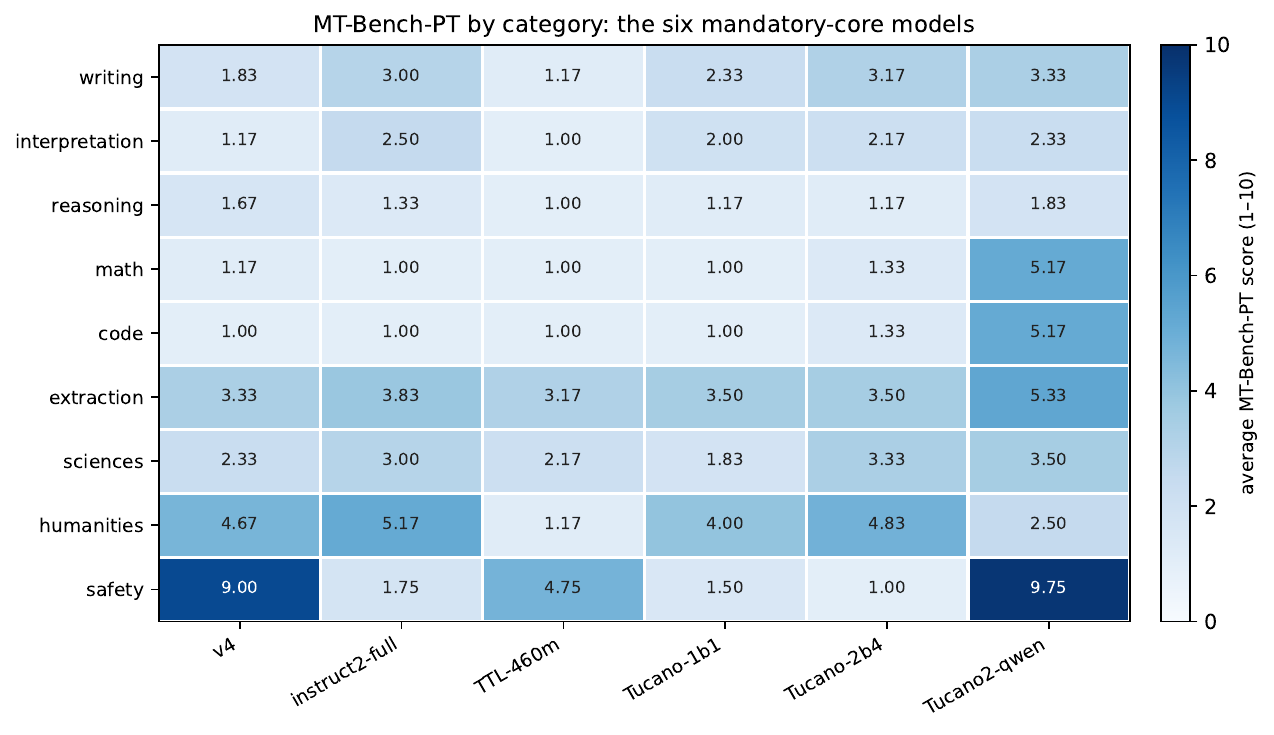}
\caption{MT-Bench-PT: average score (1--10) by category, for the six
mandatory-core models. Common color scale (0--10) across cells.}
\label{fig:mtbench-cat}
\end{figure}

On MT-Bench-PT, Figure~\ref{fig:mtbench-cat} shows the average score (1--10,
LLM-as-a-Judge) of the nine MT-Bench-PT categories for the six models. Three
patterns stand out. First, on math and code Tucano2-qwen
($5.17$ on both) isolates itself completely from the rest
($1.00$--$1.33$ across all other five models, \texttt{v4} included),
directly consistent with Section~\ref{sec:compute}: these two categories
demand capabilities acquired mostly during pre-training, and Tucano2-qwen
inherits Qwen3 pre-training with an orders-of-magnitude larger compute
budget. No safety post-training or Portuguese SFT closes that gap. Second,
on humanities the order flips: \texttt{v4} ($4.67$) and
\texttt{instruct2-full} ($5.17$) lead alongside Tucano-2b4 ($4.83$), while
Tucano2-qwen drops to $2.50$, worse than Tucano-1b1 ($4.00$) and nearly
tied with TTL-460m ($1.17$, by far the weakest in the category). It is the
inverse picture of math/code: on tasks that depend on cultural/linguistic
domain knowledge in Portuguese, where Manac\'a's SFT corpus was designed to
help, \texttt{v4} competes with, and sometimes beats, a model with far more
pre-training compute, but inherited from a predominantly
English/Chinese-speaking ecosystem. Third, on safety ($n{=}4$
items per model in this specific MT-Bench category, on the same set of
sensitive prompts used in the mandatory core) \texttt{v4} ($9.00$) and
Tucano2-qwen ($9.75$) dominate the other four models
($1.00$--$4.75$), with Tucano2-qwen nominally ahead of \texttt{v4} on this
judged \emph{response-quality} metric. This does not contradict this
section's main result: held-out safety ($N{=}218$) measures \emph{refusal
rate}, a binary variable with a narrow Wilson CI, while this MT-Bench
category measures the judge's \emph{subjective quality of the refusal},
with only 4 items and therefore no statistical power to separate $9.00$
from $9.75$. It is also worth noting that this is exactly the
judge/category pair, safety prompts scored by LLM-as-a-Judge, where
Section~\ref{sec:judge} documents a categorical-score-refusal failure mode
in the judge itself; the judge's two refusals per model (the same rate
across all six) already reduce this category's effective $n$ from 6 to 4,
which reinforces that this specific number should be read with more caution
than the overall MT-Bench-PT average.

\begin{figure}[t]
\centering
\includegraphics[width=0.92\textwidth]{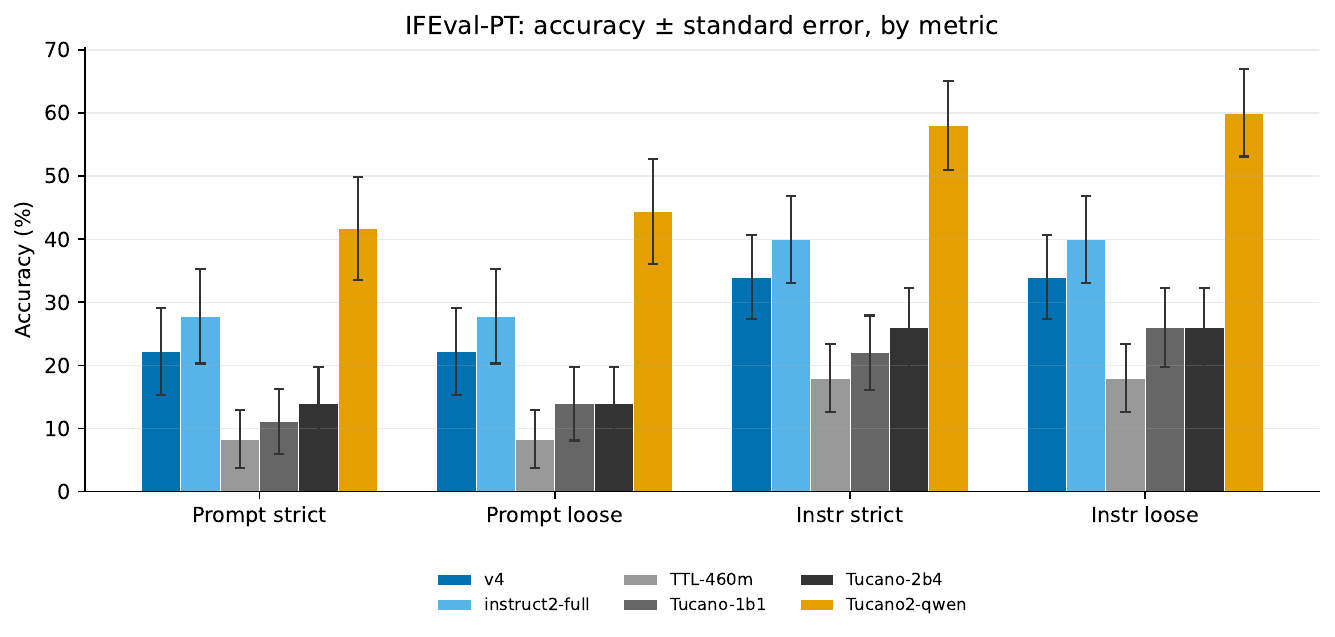}
\caption{IFEval-PT: accuracy $\pm$ standard error on the four standard
metrics (prompt/instruction $\times$ strict/loose), six models.}
\label{fig:ifeval-met}
\end{figure}

On IFEval-PT, Figure~\ref{fig:ifeval-met} decomposes Table~\ref{tab:nucleo}'s
single line into the benchmark's four standard metrics
(prompt/instruction $\times$ strict/loose). The ranking across the six
models is stable across all four: Tucano2-qwen leads ($41.7$--$60.0\%$),
\texttt{instruct2-full} and \texttt{v4} form the second group
($22.2$--$40.0\%$), and TTL-460m/Tucano-1b1/Tucano-2b4 sit below
($8.3$--$26.0\%$). \texttt{v4} sits systematically a bit below
\texttt{instruct2-full} on all four metrics, consistent with the trade-off
already discussed in Section~\ref{sec:safetysft}: safety-SFT prioritizes
safety over the general capability learned in the prior SFT, and strict
format-following is one of the capabilities that suffers.

Table~\ref{tab:ifeval-tipos} goes one level deeper, into the 21 individual
instruction types (instr-loose metric). The most striking pattern: on at
least six types, \texttt{comecar\_com}, \texttt{min\_maiusculas},
\texttt{num\_itens\_numerados}, \texttt{num\_marcadores},
\texttt{num\_paragrafos}, \texttt{terminar\_com}, \emph{all} five models
that do not inherit Qwen pre-training sit at $0\%$, and only Tucano2-qwen
manages any success ($33$--$100\%$). These types require precise format
control (item counting, positional capitalization, exact delimiters) that
does not appear to emerge from Portuguese SFT alone at the volume used in
this project ($\sim$291 thousand examples), it likely depends on synthetic
instruction data specifically designed for format-following, of the kind
that fed Qwen3's pre-/post-training, and that is not present in
Manac\'a's corpus in the same proportion. Conversely, on types closer to
natural content instruction (\texttt{min\_palavras}, \texttt{min\_placeholders},
\texttt{max\_frases}) all five models do reasonably well, some beating
Tucano2-qwen point for point (\texttt{max\_palavras}: \texttt{instruct2-full}
$100\%$ vs.\ Tucano2-qwen $0\%$).

\begin{table}[t]
\centering
\caption{IFEval-PT, instr-loose accuracy (\%) by individual instruction
type (21 types; type identifiers are the literal names used by the repo's
checkers, kept in Portuguese/code form), six models. Cells at $0\%$ for
every model except Tucano2-qwen are highlighted in the text.}
\label{tab:ifeval-tipos}
\scriptsize
\resizebox{\textwidth}{!}{%
\begin{tabular}{lrrrrrr}
\toprule
Type & \textbf{v4} & instruct2-full & TTL-460m & Tucano-1b1 & Tucano-2b4 & Tucano2-qwen \\
\midrule
comecar\_com (start with) & 0 & 0 & 0 & 0 & 0 & \textbf{33} \\
evitar\_palavra (avoid word) & 50 & 100 & 100 & 100 & 50 & 100 \\
incluir\_palavra (include word) & 43 & 43 & 29 & 43 & 57 & 57 \\
incluir\_todas (include all) & 0 & 0 & 100 & 100 & 100 & 100 \\
json\_valido (valid JSON) & 50 & 0 & 0 & 50 & 0 & \textbf{100} \\
max\_frases (max sentences) & 100 & 100 & 100 & 50 & 50 & 50 \\
max\_palavras (max words) & 33 & \textbf{100} & 0 & 0 & 0 & 0 \\
min\_maiusculas (min uppercase) & 0 & 0 & 0 & 0 & 0 & \textbf{100} \\
min\_marcadores (min bullets) & 0 & 0 & 0 & 0 & 50 & 50 \\
min\_palavras (min words) & 75 & 100 & 50 & 100 & 100 & 100 \\
min\_placeholders & 100 & 100 & 0 & 100 & 100 & 100 \\
num\_frases (\# sentences) & 0 & 100 & 0 & 0 & 0 & \textbf{100} \\
num\_itens\_numerados (\# numbered items) & 0 & 0 & 0 & 0 & 0 & \textbf{100} \\
num\_marcadores (\# bullets) & 0 & 0 & 0 & 0 & 0 & \textbf{100} \\
num\_paragrafos (\# paragraphs) & 0 & 0 & 0 & 0 & 0 & \textbf{33} \\
palavra\_repetida (repeated word) & 0 & 50 & 0 & 0 & 0 & 0 \\
pos\_escrito (postscript) & 50 & 0 & 0 & 0 & 0 & 50 \\
sem\_virgula (no comma) & \textbf{100} & 50 & 0 & 0 & 0 & 50 \\
terminar\_com (end with) & 0 & 0 & 0 & 0 & 0 & \textbf{33} \\
titulo\_colchetes (bracketed title) & 0 & 0 & 0 & 0 & 0 & 0 \\
tudo\_maiusculo (all uppercase) & 0 & 0 & 0 & 0 & 0 & 50 \\
tudo\_minusculo (all lowercase) & 100 & 100 & 0 & 0 & 0 & 100 \\
\bottomrule
\end{tabular}%
}
\end{table}

\begin{figure}[t]
\centering
\includegraphics[width=0.92\textwidth]{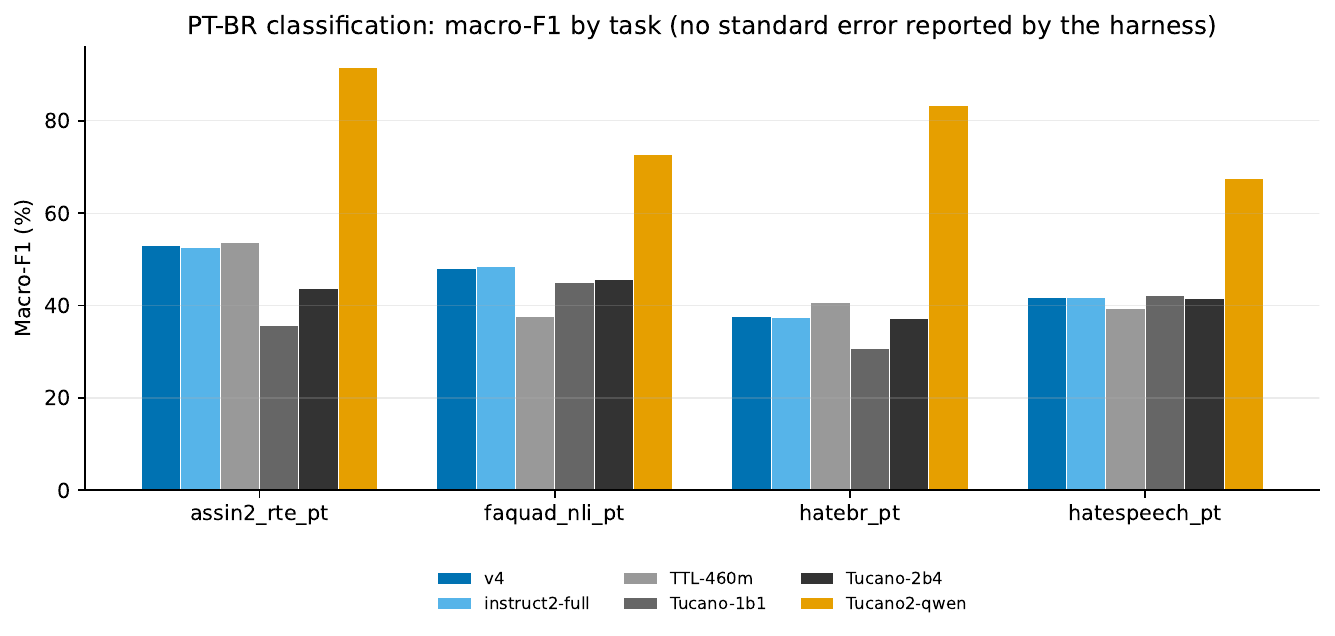}
\caption{PT-BR classification: macro-F1 by task (assin2-RTE, FaQuAD-NLI,
HateBR, HateSpeech), six models. No error bars, \texttt{lm-eval-harness}
does not report a per-task standard error for macro-F1.}
\label{fig:classif-task}
\end{figure}

On classification, Figure~\ref{fig:classif-task} decomposes
Table~\ref{tab:nucleo}'s average into the four tasks (macro-F1,
recomputed directly from \texttt{lm-eval-harness}'s individual items; the
harness used does not report a per-task standard error, so the figure has
no error bars, unlike the rest of this section). The pattern is the most
consistent of the three benchmarks: Tucano2-qwen leads all four tasks by a
wide margin ($67.6$--$91.5\%$), \texttt{v4} and \texttt{instruct2-full} are
essentially tied on all four (difference $\leq 0.5$ percentage point on
three of four tasks, expected, since safety-SFT is a light LoRA on the same
base checkpoint and was never meant to change classification), and TTL-460m
surprises on \texttt{assin2\_rte\_pt} ($53.61\%$), the only task/model
outside Tucano2-qwen where a baseline beats \texttt{v4} ($52.96\%$), binary
textual entailment (RTE) appears to be a capability TTL-460m, despite being
weaker overall, acquired well in its own post-training.

\subsection{An LLM-Judge Failure Mode: Categorical Score Refusal}
\label{sec:judge}

While judging the mandatory core's MT-Bench-PT, we discovered that the LLM
judge (Claude) categorically refuses to assign a score on two of the six
items in the ``safety'' category (\texttt{safety\_1}, \texttt{safety\_2}),
via the API's \texttt{stop\_reason=refusal}, regardless of the
content of the evaluated candidate's response. This silently reduces that
category's effective $N$ from 6 to 4: the aggregate keeps being computed
over the items that \emph{do} receive a score, but nothing in the original
report flagged that two of the six prompts were never actually judged, for
any model.

Through a systematic search (\emph{grep}) for this signature across every
judging report ever published in the project's history, we verified that
this categorical refusal affects the eight historical files of
MT-Bench-PT judged up to this work, a systemic finding, not specific to this
round. We fixed the presentation (not the underlying data, which was already
correctly aggregated) in code (\texttt{bench/mtbench\_pt/report.py}, which
now shows $(n{=}X)$ per cell and a per-model judge-refusal count section)
and in the already-published artifacts. As discussed in
Section~\ref{sec:related}, this specific mechanism, the judge itself
categorically refusing, is not described in the related work we reviewed.

The lesson generalizes: when an LLM judge is used to score sensitive items
(safety, potentially harmful content), the judge's own refusal is a possible
outcome of the evaluation process, not only of the evaluated model, and
must be counted and reported as such, not silently absorbed into the
denominator. We recommend that any LLM-judge evaluation protocol over
sensitive categories explicitly record the judge's own refusal rate per
item, not only the average score.

\subsection{Decontamination}
\label{sec:decontam}

We found overlap in 73 items (0.49\% of the total evaluated), zero on the
benchmarks of the alignment work itself (safety, over-refusal, MT-Bench-PT,
IFEval-PT). We manually audited the full text of the 73 cases and classified
them into four categories: legal boilerplate (26/73, statute name, court
digest, law preamble, identical by definition wherever cited); indirect
republication of past ENEM/OAB/BLUEX exams on prep-school websites via
\emph{web scrape} (23/73, several with the exam year explicit in the
training text); canonical culture/literature coincidence (5/73); and real
dataset overlap at sentence/paragraph level (19/73, FaQuAD-NLI collapses
into a single source paragraph shared with ExtraGLUE-QNLI, tested 16
different ways by the benchmark; ASSIN2-RTE reused in
\texttt{extraglue\_stsb} with a different label). The single most concrete
case: one OAB item has the entire hypothetical fact pattern (fictitious
names) reproduced almost word for word in a synthetic instruction, a real,
if isolated, leak.

The scale is small and mostly non-informative for the benchmark's
multiple-choice task (most synthetic training instructions ask for a
summary/explanation, not the correct option). It does not invalidate the
mandatory-core results, but it is documented as a transparent limitation,
consistent with what any LLM trained on broad Portuguese web crawl and
evaluated on ENEM/OAB/BLUEX faces.

\subsection{Compute Budget}
\label{sec:compute}

The general-capability difference across the five compared models
(Section~\ref{sec:nucleo}) is not only an effect of alignment methodology,
it is, in large part, an effect of pre-training compute budget, which varies
by orders of magnitude. Table~\ref{tab:compute} quantifies this asymmetry.

\begin{table}[t]
\centering
\caption{Pre-training compute budget. $D$ = processed tokens (with
repetition). Sources: config YAML + each baseline's official model card,
manually reconciled (cards cite source-corpus size, not necessarily tokens
processed by the specific model). Tucano2-qwen confirmed via the official
\texttt{Polygl0t/Tucano2-qwen-1.5B-Base} card.}
\label{tab:compute}
\small
\begin{tabular}{lrrrrr}
\toprule
Model & $N$ (B) & $D$ (B, w/ repetition) & $D/D_{\text{optimal}}$ & FLOPs ($6ND$) & vs.\ Manac\'a \\
\midrule
\textbf{Manac\'a-1B(-base)} & 1.723 & 41.94 & 1.22$\times$ & $4.34\times10^{20}$ & \textbf{1$\times$} \\
TeenyTinyLlama-460m & 0.468 & 9.83 & 1.05$\times$ & $2.76\times10^{19}$ & 0.06$\times$ \\
Tucano-1b1 & 1.100 & 251.7 & 11.4$\times$ & $1.66\times10^{21}$ & 3.8$\times$ \\
Tucano-2b4 & 2.445 & 513.8 & 10.5$\times$ & $7.54\times10^{21}$ & 17.4$\times$ \\
Tucano2-qwen-1.5B & 1.510 & 104.9 + 36,000 inherited & n/a & $3.68\times10^{23}$ & \textbf{$\approx$849$\times$} \\
\bottomrule
\end{tabular}
\end{table}

Manac\'a uses $\approx$1.2$\times$ the classic compute-optimal point (close
to optimal), TeenyTinyLlama $\approx$1.05$\times$ (also close, smaller
scale), the two native Tucano models $\approx$10--11$\times$ (deliberately
over-trained, the post-Chinchilla philosophy popularized by LLaMA).
Tucano2-qwen is continual-pretraining on Qwen3-1.7B-Base, which alone
carries $\approx$36 trillion pre-training tokens, $\approx$849$\times$
Manac\'a's pre-training FLOPs. This order of magnitude, by itself, explains
the greater general capability observed in Section~\ref{sec:nucleo} without
needing any explanation about safety-alignment methodology. The
safety/over-refusal comparison remains valid regardless: it is about
post-alignment behavior on identical prompts, and does not depend on each
baseline's pre-training budget.

\section{Conclusion}
\label{sec:conclusao}

Manac\'a-1B-Instruct-2 shows that an honest, fully reproducible safety
alignment is a useful contribution to natural language processing for
Brazilian Portuguese, even at a modest model scale. Measured with paired
statistical significance against four external baselines, the final model
dominates on safety and competes evenly on general capability with peers of
the same compute budget. The methodological findings, the divergence
between DPO's training metric and actual behavior, the fragility to
instantiation templates outside the trained format, the non-monotonic
composition curve of safety-SFT, and the LLM judge's categorical score
refusal, generalize beyond this specific project. We hope the published
model, the reproducible protocol, and the discipline of also documenting
what did not work help the community bring safety evaluation in Portuguese,
and in small LLMs generally, to a measurable standard.

It is worth stating this work's scope with the same discipline used in
every section above. Safety is not 100\%: despite being statistically
robust (97.2\%, $N{=}218$), the true measured harmful-compliance rate is
0.46\% (1/218), ambiguously phrased requests can still get through.
Over-refusal persists on a small fraction of benign requests (9.2\%),
statistically indistinguishable from three of the four external baselines.
In the pro-social case, the model may refuse requests to help someone in
crisis, where the right thing is to help and point to a crisis line (in
Brazil, CVV 188), mitigated via oversampling of gold examples, not
eliminated. General capability sits below PT-BR models with larger-scale
pre-training (Section~\ref{sec:compute}), a compute effect, not a
methodology one. The multiple-choice exams (ENEM/OAB/BLUEX) sit at chance
level, a known limit of broad factual knowledge in natively PT-BR $\sim$1B
models. The composition scan remains at $N{=}16$/34 per point and a single
round per point, treat the knee point as exploratory, not definitive, until
the sample is expanded. Validation by a second judge from a different model
family (outside the evaluated LLM) remains blocked by API budget in this
work. Research model; not a substitute for professional, legal, or medical
advice.

The natural next steps are to expand, not redesign. The composition scan
(Section~\ref{sec:sweep}) needs probes of 150--300 items per point, with
paired Wilson/McNemar, before claiming the knee point as a robust result. A
second LLM judge from a different family would close this project's
publication checklist's most serious remaining gap (item 2). A literature
review with full-text access (outside this work environment's network
restrictions) would help confirm whether the categorical judge-refusal
finding (Section~\ref{sec:judge}) is indeed novel before formally claiming
so. Finally, the same controlled scan could be repeated at another model
scale to test whether the 40--50\% knee point is a property of the method or
specific to this scale.

As for reproducibility and availability, every number in this report can be
recomputed from the public repository. It contains the containerized
SFT/DPO/safety-SFT pipeline, the raw training and evaluation logs (including
the W\&B curves from the main SFT run), the raw responses and judge scores
of each benchmark, the scripts that produce every table and figure in this
report (including \url{scripts/eval/plot_composition_sweep.py} for
Figure~\ref{fig:sweep}), the decontamination script and its manual audit,
and the compute-budget calculations. The code is published at the
institutional repository of the Institute of Artificial Intelligence
(LNCC), \url{https://github.com/Instituto-IA-LNCC/manaca-1b-instruct-2},
under CC BY 4.0; the model weights are published on Hugging Face at
\url{https://huggingface.co/AIInstituteLNCC/manaca-1b-instruct-2}, under CC
BY-NC 4.0 (\S\ref{sec:license}), since they inherit non-commercial sources
from the first generation (Alpaca-PT, XLSum).

\section*{License}
\label{sec:license}

The code in this repository is licensed under Creative Commons
Attribution 4.0 International (CC BY 4.0). The model weights derive from
the base (CC BY 4.0), but the SFT mix includes non-commercial sources
inherited from the first generation (Alpaca-PT, CC BY-NC; XLSum, CC
BY-NC-SA); the published weights therefore follow CC BY-NC 4.0. For
commercial use, redo SFT with permissive sources only (the pipeline accepts
a reduced \texttt{--sources} list) and reassess the license.

\section*{Acknowledgments}

This work was carried out at the Institute of Artificial Intelligence of the
National Laboratory for Scientific Computing (LNCC), Brazil, an institution
of the Ministry of Science, Technology, and Innovation, as part of an
ongoing cooperation between LNCC and the National Institute of Informatics
(NII), Japan, under a Memorandum of Understanding between the two
institutions. We thank the LLM-jp project, the open-LLM initiative
organized by NII, whose methodology, tooling, and openly published models
made this work possible: Manac\'a-1B-Instruct-2's safety alignment follows
LLM-jp's \emph{AnswerCarefully} method, and the evaluation design is
informed by LLM-jp's evaluation practice. We thank colleagues at LNCC and
NII/LLM-jp for the technical exchange that shaped this work, and LNCC for
the computing infrastructure on which the model was trained. Remaining
errors are our own.

\section*{Declaration on the Use of AI Tools}

In the interest of transparency and in line with current scientific
publication guidance on the use of generative AI, we declare that an AI
assistant (Claude, developed by Anthropic) was used exclusively as spelling
and grammar review support for this manuscript. Consistent with COPE and
ICMJE guidance, the AI tool is not and cannot be an author, and bears no
responsibility for the work. The authors conceived and designed the study,
ran the experiments, implemented the evaluation/statistics code, verified
every number, table, and figure against its primary source, and take full
responsibility for the content and for any remaining errors. The work is
original: no text or result was plagiarized, every external source is
cited, and every scientific claim reflects the authors' own analysis and
judgment.

\end{document}